\documentclass[10pt,twocolumn,letterpaper]{article}

\usepackage{conference}              %

\usepackage{amsmath,amssymb,amsfonts}
\usepackage{algorithmic}
\usepackage{graphicx}
\usepackage{textcomp}
\usepackage{xcolor}
\usepackage{booktabs}
\usepackage{makecell}
\usepackage{tikz}
\usetikzlibrary{positioning, calc, fit, patterns, shapes, backgrounds, external}
\usepackage[acronym]{glossaries}
\usepackage{algorithm}

\newtheorem{definition}{Definition}

\newacronym{MDP}{MDP}{Markov decision process}
\newacronym{SAC}{SAC}{soft actor-critic}
\newacronym{RL}{RL}{reinforcement learning}
\newacronym{RIBA}{RIBA}{reinforcement learning inspired black-box adversarial attack}
\newacronym{ASR}{ASR}{attack success rate}

\definecolor{conferenceblue}{rgb}{0.21,0.49,0.74}
\usepackage[pagebackref,breaklinks,colorlinks,allcolors=conferenceblue]{hyperref}

\title{Reinforcement Learning Inspired Black-box Adversarial Attacks for Computer Vision}

\author{Florian Krone \and Elena Hoemann \and Sven Hallerbach\and
Institute for AI Safety and Security\\
German Aerospace Center (DLR)\\
Sankt Augustin, Germany \\
{\tt\small \{florian.krone, elena.hoemann, sven.hallerbach\}@dlr.de}
}

\begin{document}
\crefname{definition}{Def.}{Defs.}
\Crefname{definition}{Definition}{Definitions}
\crefname{algorithm}{Alg.}{Algs.}
\Crefname{algorithm}{Algorithm}{Algorithms}
\twocolumn[{
\maketitle
    \begin{center}
        \includegraphics{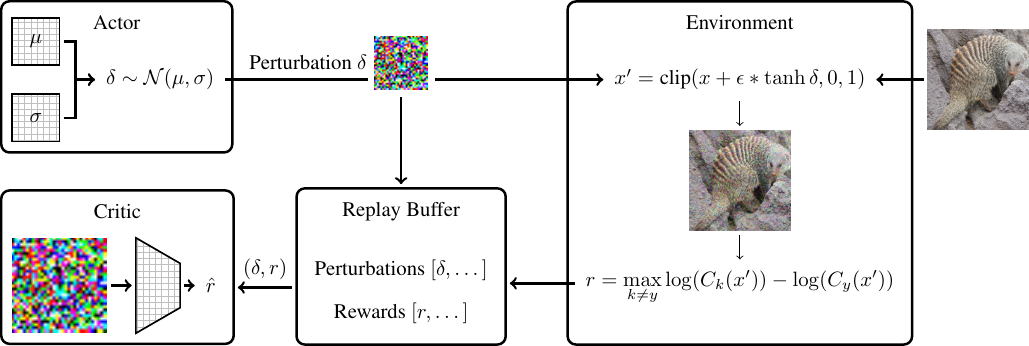}

        \captionof{figure}{Visualization of the \acrshort{RIBA} algorithm. The algorithm is inspired by a \acrlong{SAC}, but tailored for the specific case of generating adversarial perturbations. A perturbation is generated by the actor, using the reparameterization trick. The actor is trained against a critic, which learns to evaluate adversarial perturbations.}
        \label{fig:rl-for_attacks}
    \end{center}
}]
\begin{abstract}
Neural networks, both convolution or transformer based, are essential for modern computer vision systems. However, they are vulnerable to small perturbations, almost imperceptible to humans, which significantly alter the model's prediction. These adversarial attacks are often considered to be a significant threat to the implementation of neural networks in safety-critical applications. Most attacks utilize the white-box threat model and therefore require full access to the target model, making them unrealistic to use in practice. We propose a novel approach under the more realistic black-box threat model that utilizes concepts from reinforcement learning to optimize perturbations with a non-differentiable target model. Reinforcement learning algorithms have already been optimized to be query efficient, making them an ideal starting point when designing black-box adversarial attacks. We show the success of our reinforcement learning inspired black-box adversarial attack (RIBA) in generating adversarial perturbations using only a small number of queries to the target model, by comparing it to state of the art attacks on different models on the Cifar10 and ImageNet data sets. RIBA takes $25.4\%$ fewer median queries to generate attacked images against a ResNet-18 on Cifar10 and $22.5\%$ fewer median queries to fool a Vit-B/16 model on ImageNet. Additionally, we demonstrate that RIBA can match the performance of white-box attacks on an adversarially trained model.
\end{abstract}

\section{Introduction}
\label{sec:introduction}

Neural networks have shown remarkable success in computer vision tasks and are therefore increasingly applied in the real world, including safety-critical systems, such as automated vehicles. However, it was shown that they are susceptible to adversarial attacks \citep{Szegedy2014}, \ie, perturbations to the input image that are insusceptible to humans. Adversarial attacks can be categorized by the information they require about the targeted model. The white-box threat model requires full access to the model and its weights, making it possible to utilize gradient information in calculating the perturbations. The black-box model, in contrast, treats the model as an oracle, where only the predictions for inputs are available. Two sub categories of the black-box model exist, where the attacker knows either just the final decision of the model or the scores of all classes, as is the case with our attack. Attacks designed for the black-box threat model have a more widespread use, as they for example can be used to attack AI models deployed behind a web API. The web API use case also motivates an important secondary goal of black-box attacks. Since every query to the API is typically associated with a cost, black-box attacks aim to find perturbations with as little queries as possible. To quantify if an attack is insusceptible to a human, a norm bound is applied to the perturbation, limiting the difference between the benign and the attacked sample. Our approach utilizes the \(l_\infty\)-norm, which is commonly used to limit the perturbation \citep{Goodfellow2015, Ilyas2018}.

The main contributions of this paper can be summarized as follows:
\begin{itemize}
    \item We formulate the problem of generating adversarial perturbations as a \gls*{MDP}. This allows for the application of a \acrlong{RL} algorithm to generate adversarial perturbations.
    \item We propose a novel algorithm to solve this specific \gls*{MDP} based on \acrfull{SAC} \citep{Haarnoja2018}.
    \item The success of our novel approach is demonstrated on standard benchmarks and compared to state of the art attacks.
\end{itemize}
To highlight the close relation of our approach to \gls*{RL}, we name it \gls*{RIBA}.

\section{Related work}
\label{sec:related-work}

Black-box adversarial attacks, decision or score based, can be built in a number of different ways. In the following section, we introduce approaches that are gradient, random-search, or transfer based, as well as other \gls*{RL} based approaches. 

A common approach is to utilize concepts from white-box attacks like PGD \citep{Madry2018a} or C\&W \citep{Carlini2017}. However, these attacks require access to the gradient to optimize the perturbations. Black-box attacks, therefore, need to estimate the gradient first. Notable attempts are ZOO \citep{Chen2017}, ZO-NGD \citep{Zhao2020}, ZO-AdaMM \citep{Chen2019} and NES \citep{Ilyas2018}. In the Bandits attack \citep{Ilyas2019}, the query count to the attacked model was reduced by integrating prior knowledge about the gradients into the estimation. Specifically, a time-dependent prior, reusing knowledge from previous steps of the attack optimization and a data-dependent prior, using local similarity within images, were introduced.

A different class of attacks utilizes random search to optimize adversarial perturbations. These attacks are specifically built for the black-box threat model. \citet{Guo2019} introduced a random-search based attack called SimBA that randomly samples a vector from an orthonormal base and adds or subtracts it to the image.
Another random-search based attack was introduced by \citet{Andriushchenko2020}, who iteratively add squares of different sizes and colors to the image until the target model is fooled. Consequently, their attack is called Square Attack.
\citet{Bai2023} propose to reduce the search space by utilizing an encoder/decoder setup in their NP-Attack. The original image is first encoded into a low dimensional space, where a random perturbation is sampled and added to the image. The perturbed image is then decoded and it is tested whether the perturbation fools the target model.

Adversarial attacks are known to transfer between models to a certain degree, \ie, a perturbation generated for an image against one model might also fool a different model on the same image. This property can be exploited to attack a black-box model, utilizing a white-box surrogate model and transferring the perturbation. One example is the EigenBA attack \citep{Zhou2022}. \citet{YANG2020} additionally query the targeted model in the LeBA attack to update the surrogate model continuously. The Simulator attack \citep{Ma2021} works in a similar manner, by training and continuously updating a model that simulates the response of the target model. To improve the generalization, the SVRE \citep{Xiong2022} attack uses an ensemble of surrogate models.

A few approaches have applied \gls*{RL} before to generate adversarial examples. \citet{Huang2023} proposed their DBAR attack. They formulate the decision-based black-box adversarial attack problem as an \gls*{MDP} and utilize an \acrshort{RL} algorithm to solve it. \citet{Kang2023} additionally employ an autoencoder to reduce the dimensionality of the search space. However, these approaches utilize general \gls*{RL} algorithms to solve the problem, that disregard problem specific knowledge, which could be used for further optimizations. With this work, we aim to address this issue. 
A different paradigm of \gls*{RL}-based attacks exists, in which an agent is first trained in a dedicated training phase and than later applied to generate perturbations. Examples are RLAB \citep{Sarkar2023a} and QTRL \citep{Ma2025}. However, this setup differs from the typical goal of black-box attacks, which usually aims to generate perturbations independent of each other, with as few queries as possible. Our approach is in line with the typical goal of black-box attacks.

\section{Method}
\label{sec:method}
The main contribution of this paper is a novel approach to build adversarial perturbations using a method inspired by \gls*{RL}, while requiring only black-box access to the model. The application of \gls*{RL} requires the problem of optimizing an adversarial perturbation to be formulated as an \acrshort{MDP}, which is outlined in the following section. \Cref{sec:adapted-rl} describes the \gls*{RL} algorithm we used and the problem specific changes we made to it.
\subsection{Adversarial attacks as a MDP}
\label{sec:adversarial-mdp}
\begin{definition}
    \label{def:adversarial-attack}
    Given a classifier \(C\colon [0,1]^{H\times W\times 3}\rightarrow\mathbb{R}^N\) and a sample, label pair \(x\in[0,1]^{H\times W\times 3}, y\in\mathbb{R}^N\) with \(C(x)=y\), then \(\delta \in \mathbb{R}^{H\times W\times 3}\) is an adversarial perturbation, if \(C(x+\delta)\neq y\) and \(\lVert\delta\rVert_\infty<\epsilon\) for a budget \(\epsilon\).
\end{definition}
\begin{definition}
    \label{def:mdp}
    Given a set of states \emph{S}, 
    a set of actions \emph{A}, 
    a transition probability function \(P\colon S\times A\times S \rightarrow [0,1]\), 
    and a reward function \(R\colon S\times A\times S \rightarrow \mathbb{R}\),
    the tuple \((S, A, P, R)\) is called a \acrfull{MDP}.
\end{definition}
The transition function \(P\colon S\times A\times S \rightarrow [0,1]\) denotes the probability to reach state \(s_{i+1}\in S\), if action \(a\in A\) is taken in state \(s_i\in S\). Formally, it is defined as
\begin{align}
    P(s_i, a, s_{i+1}) = \Pr(s_{i+1}\mid s_i, a).
\end{align}
Consequently, the reward function \(R\colon S\times A\times S \rightarrow \mathbb{R}\) describes the reward obtained for taking action \(a\in A\) in state \(s_i\in S\) and reaching state \(s_{i+1}\in S\). The goal of reinforcement learning is to optimize an agent that, given a sate \emph{s}, predicts the optimal next action \emph{a}, so that the reward over a series of state-action-state triplets, called an episode, is maximized. Connecting this to \cref{def:adversarial-attack}, we want to train an agent that, given a sample \emph{x}, creates an adversarial perturbation \(\delta\).
The state is given by the current image \(x\in S\), where \(S=[0,1]^{H\times W\times 3}\). 
The actions are given by the adversarial perturbations. We define a deterministic transition function, where, for an action \(\delta\), the state does not actually change, \ie, we are still considering the same image. 
The formal definition is given by
\begin{align}
P(x_i, ~\delta, x_{i+1}) &= \Pr(x_{i+1}\mid x_i, \delta)\\
\label{eq:mdp-transition}
&= \begin{cases}
    1 & \text{if } x_{i+1} = x_i,\\
    0 & \text{otherwise.}
\end{cases}
\end{align}
Finally, the reward function depends on the model we are attacking. Considering a classification model \(C\colon[0,1]^{H\times W\times 3}\rightarrow[0,1]^N\) with \emph{N} classes, we define \(C_k(x)\) as the probability of the \(k\)-th class, for a given \(x\in[0,1]^{H\times W\times 3}\). Additionally, we define a function \(p\) to apply a perturbation to an image.
\begin{align}
    \label{eq:apply-perturbation}
    p(x, \delta)=\text{clip}(x+ \epsilon*\tanh \delta, 0, 1).
\end{align}
The \emph{clip} functions limits the output to the valid range of images, \ie, each pixel in \([0,1\)], while the term \(\epsilon * \tanh \delta\) ensures that the perturbed images are within the \(l_\infty\)-norm bound around the original image. A perturbation \(\hat{\delta}\) according to \cref{def:adversarial-attack} is obtained by \(\hat{\delta} = p(x,\delta)-x\).
The reward for an action \(\delta\in A\) in state \(x_i\in S\) that leads to the next state \(x_{i+1} \in S\) is then given by
\begin{align}
    R(x_i, &\delta, x_{i+1}) \nonumber\\
    \label{eq:reward}
    &= \underset{k\neq y}{\max}\log (C_k(p(x_i,\delta))) - \log (C_y(p(x_i,\delta))),
\end{align}
where \emph{y} is the ground truth of the attacked image. The reward function is inspired by the loss function used in \citep{Chen2017} to generate adversarial perturbations. We removed the parameter \(\kappa\), which was originally used to control the strength of the perturbation. We are aiming to find a perturbation with as little queries as possible and consequently take the first perturbation that is found. The parameter \(\kappa\) would therefore not have any effect. The \(\log\) function helps to emphasize small changes to the model's output made by the perturbations, therefore boosting the sample efficiency. With this setup, it is already possible to generate adversarial perturbations. The next section describes the specific setup we used, and the changes we made to the \gls*{RL} algorithm to better suit our specific problem.
\subsection{Specific RL algorithm}
\label{sec:adapted-rl}
The setup in \cref{sec:adversarial-mdp} can be used to train an \acrshort{RL} agent that is able to generate adversarial perturbations for different inputs. However, our goal was to build a typical black-box adversarial attack that does not require any pretraining, but calculates a perturbation for an individual image based on a limited number of queries to the targeted model. Therefore, we train an individual \gls*{RL} agent for each image and terminate the training, as soon as a perturbation is found. Additionally, we terminate each episode after one step, as according to \cref{eq:mdp-transition}, the state does not change.
To train the \gls*{RL} agents, we use a modified version of \gls*{SAC} \citep{Haarnoja2018}. One of the key concepts of off-policy \gls*{RL}, like \gls*{SAC}, is to separate the collection of experience from the training, using a replay buffer. First, a number of actions are executed in the environment to gain experience. This is stored in the replay buffer. Afterwards, batches from the replay buffer are sampled for training. Note that the training is not necessarily done on the most recent experience.
We train a critic \(Q_\theta(\delta)\) that estimates the expected reward for a perturbation \(\delta\). The critic is trained using 
\begin{align}
\label{eq:critic-objective}
    J_Q(\theta) = \mathbb{E}_{(\delta,r) \sim D}[(Q_\theta(\delta)-r)^2],
\end{align}
where \emph{D} represents the replay buffer. Known perturbation, reward pairs are sampled from the replay buffer and the critic is trained to estimate the reward. The simplicity of this task allows us to use a very small network for the critic, consisting of a fully connected network with a singe hidden layer.
The second part of our algorithm is the actor, which contains a policy \(\pi_\phi\) used to generate perturbations. The policy utilizes the reparameterization trick, it consists of two sets of learned parameters \(\mu_\phi\) and \(\sigma_\phi\) used to parameterize a normal distribution. The perturbations are then sampled from the distribution.
The training objective for the policy is given by
\begin{align}
    \label{eq:policy-objective}
    J_\pi(\phi) = \mathbb{E}_{\xi\sim\mathcal{N}}[\alpha\log\pi_\phi(f_\phi(\xi))-Q_\theta(f_\phi(\xi))].
\end{align}
The objective uses the reparameterization trick with 
\begin{align}
    \label{eq:reparametrization}
    f_\phi(\xi) = \mu_\phi + \sigma_\phi\xi,~ \xi\sim\mathcal{N}(0,1).
\end{align}
The training objective for the policy contains two separate terms. The term \(Q_\theta(f_\phi(\xi))\) trains the policy to maximize the reward, using the critic as an estimator for the reward. Using the critic here is a key part of \gls*{SAC}, as this allows to train the policy without directly interacting with the environment, \ie, querying the target model in our case. The term \(\log\pi_\phi(f_\phi(\xi))\) forces the policy to have a high entropy. The parameter \(\alpha\) is used to balance the two objectives. We do not set \(\alpha\) directly, but optimize it as introduced in \citep{Haarnoja2019}.
\begin{algorithm}[tb]
    \caption{RIBA}
    \label{alg:riba}
    \begin{algorithmic}[1]
        \REQUIRE sample-label pair \(x,y\)
        \ENSURE adversarial perturbation \(\delta\)
        \STATE initialize \(\pi_\phi\) and \(Q_\theta\)
        \STATE \(D \gets \emptyset\)
        \FOR{each iteration}
            \STATE \(\delta \sim \pi_\phi\)
            \label{alg-line:sample-perturbation-start}
            \STATE \(r \gets \underset{k\neq y}{\max}\log(C_k(p(x, \delta))) - \log(C_y(p(x, \delta)))\)
            \IF{\(\underset{k}{\text{argmax }}C_k(p(x,\delta)) \neq y \)}
                \RETURN \(\delta\)
            \ENDIF
            \STATE \(D \gets D \cup \{(\delta, r)\}\)
            \label{alg-line:sample-perturbation-end}
            \FOR{each critic update step}
            \label{alg-line:training-start}
                \STATE \(\theta \gets \theta - \eta\nabla_\theta J_Q(\theta)\)
            \ENDFOR
            \FOR{each actor update step}
                \STATE \(\phi \gets \phi - \eta\nabla_\phi J_\pi(\phi)\)
            \ENDFOR
            \label{alg-line:training-end}
        \ENDFOR
    \end{algorithmic}
\end{algorithm}
\Cref{alg:riba} provides an overview, how we generate adversarial perturbations. Just like in \gls*{SAC}, our algorithms alternates between collecting experience in the environment, which is then stored in the replay buffer and updating the critic and policy. In our case, collecting experience in the environment requires calculating the reward using \cref{eq:reward} and checking if a perturbation was found. The perturbation, reward tuple \((\delta,r)\) is then stored in the replay buffer. Afterwards, the critic and policy are updated using \cref{eq:critic-objective} and \cref{eq:policy-objective} respectively with a learning rate \(\eta\). 
To increase the sample efficiency, the critic and policy are each updated multiple times before a new perturbation is tried against the target model. \Cref{fig:rl-for_attacks} provides a visualization of the algorithm.

While our \gls*{RIBA} algorithm is inspired by \gls*{SAC}, a number of modifications have been made to better suit the specific problem at hand. Most importantly, we removed the dependence on the state from both the critic and the policy.
Typically, the policy is a conditional distribution \(\pi_\phi(\cdot\mid s)\) over the actions, given the current state \(s\). In this case, both \(\mu_\phi(s)\) and \(\sigma_\phi(s)\) are neural networks predicting the parameters for the normal distribution based on the state.
The critic \(Q_\theta(s,a)\) also depends on the state, estimating the expected return if action \(a\) is executed in state \(s\). The return of an episode describes the discounted sum of all rewards in that episode. In our case, each episode is exactly one step long. Therefore, the return is given by the reward for that step. This is an import observation, as it removes the need for target Q-functions to train the critic, which simplifies the training objective.

\section{Experiments}
\label{sec:experiments}
To evaluate the effectiveness and query efficiency of our approach in generating adversarial perturbations against black-box models, we attack models trained on Cifar10~\citep{Krizhevsky2009} and ImageNet \citep{Russakovsky2015}. For the evaluation on Cifar10, we use a ResNet-18 \citep{He2016} model. On ImageNet, we use an Inception-v3 \citep{Szegedy2016} and a ViT-B/16 \citep{Dosovitskiy2021} model to investigate the differences between attacking convolutional and transformer based architectures. \Cref{tab:hyperparameters} shows the experimental setup and hyperparameters. All hyperparameters are optimized targeting a ResNet-18 model on Cifar10 and used across all experiments. To generate a perturbation of size \(32\times32\times3\), the critic is a standard MLP \(Q:\mathbb{R}^{3072}\rightarrow \mathbb{R}\) with a single hidden layer of size \(6\). The actor consists of a tuple \(\mathbb{R}^{3072}\times \mathbb{R}^{3072}\) which is used to generate actions according to \cref{eq:reparametrization}. We utilize the AdamW \citep{Loshchilov2019} optimizer with a learning rate of \(0.02\). The learning rate is decayed by a factor of \(0.1\) if the reward hits a plateau. To boost the sample efficiency, both the actor and the critic, are updated multiple times per query to the target model. This is one of the core functionalities of off-policy \gls*{RL} and the main reason for the query efficiency of our attack. In particular, we update the actor \(120\) times and the critic \(40\) times per query to the target model. To boost the initial exploration, we initialize the replay buffer with \(5\) randomly sampled perturbations. The amount of samples is a trade-off between improved exploration and an increase in queries, as each sample comes at the cost of one query to the target model. The trade-off between the entropy of the policy and maximizing the reward in \cref{eq:policy-objective} is optimized according to \citep{Haarnoja2019}. Typically, the target entropy is set to \(-n\), where \(n\) is the number of dimensions in the action space. In our case this would be \(-3072\), far greater than for example in a typical robotic application. We found using the number of channels, \ie, \(-3\), to lead to better results.
\begin{table}[tb]
    \centering
    \caption{Experimental setup and hyper parameters}
    \label{tab:hyperparameters}
    \begin{tabular}{llr}
    \toprule
    Parameter & Description & Value \\
    \midrule
    max queries & Maximum allowed queries & 10000\\
    \(\epsilon\) & \makecell[l]{Perturbation budget\\ under \(l_\infty\) norm} & \makecell[r]{0.031\\/ 0.05}\\ 
    \makecell[l]{critic\\ update steps} & \makecell[l]{Number of critic updates\\ per query} & 40\\
    \makecell[l]{actor\\ update steps} & \makecell[l]{Number of actor updates\\ per query} & 120\\
    \(\eta\) & Learning rate & 0.02\\
    start samples & \makecell[l]{Add random samples to the\\replay buffer before \cref{alg:riba}} & 5\\
    initial \(\alpha\) & parameter in \cref{eq:policy-objective} & 0.1\\
    target entropy & \makecell[l]{Optimize \(\alpha\)\\according to \citep{Haarnoja2019}} & -3\\
    \makecell[l]{critic hidden\\ dimension} & \makecell[l]{Single hidden layer\\ of the critic} & 6\\
    critic batch size & Batch size for critic updates & 400\\
    actor batch size & Batch size for actor updates & 650\\
    \bottomrule
\end{tabular}

\end{table}

We compare our approach to a number of state of the art black-box attacks that do not require any additional information such as surrogate models or pretraining against the target model. We use the open sourced implementations and suggested hyperparameters of the attacks to evaluate them against the same models and images we used for our attack. As it is common practice, the maximum amount of queries is limited to \(10000\) and an attack is considered as failed if no perturbation was found in this limit. The reported average and median amount of queries required are only calculated for the successful attacks.
\subsection{Cifar10 Evaluation}
For the evaluation of our attack on Cifar10, we use a ResNet-18 model with a benign accuracy of \(94.98\%\). We limit the perturbations to the typical budget of \(\epsilon=8/255\) and evaluate the attacks on all \(9498\) correctly classified images of the test set. \Cref{tab:result-cifar10} shows the \gls*{ASR}, as well as the average and median amount of queries required to generate a perturbation that fools the model.
\begin{table}[tb]
    \centering
    \caption{Experimental evaluation of attacking a ResNet-18 model trained on Cifar10 with a perturbation limit of \(8/255\) under the \(l_\infty\) norm.}
    \label{tab:result-cifar10}
    \begin{tabular}{lrrr}
    \toprule
    Attack & ASR & \makecell[r]{Avg.\\Queries} & \makecell[r]{Median\\Queries} \\
    \midrule
    SimBA & 99.81\% & 233 & 116 \\
    Square Attack & \textbf{100\%} & 147 & 67 \\
    RIBA (ours) & \textbf{100\%} & \textbf{107} & \textbf{50} \\
    \bottomrule
\end{tabular}

\end{table}
It was already possible to reliably generate perturbations under this setup with the Square Attack. However, our \gls*{RIBA} algorithm requires significantly less queries to the target model, both on average and median.
\begin{figure}[tb]
    \centering
    \includegraphics[width=.47\textwidth]{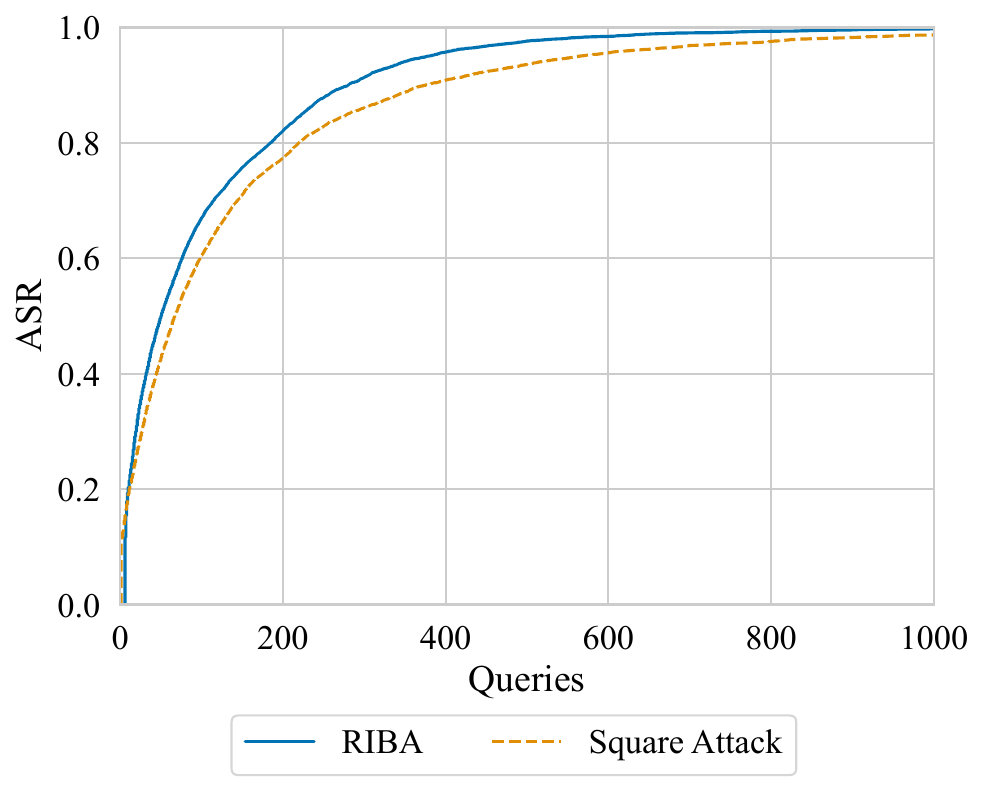}
    \caption{Comparison of the attack success rate of \gls*{RIBA} and Square Attack at different query limits when attacking a ResNet-18 model on Cifar10.}
    \label{fig:riba-square-comparison-cifar10}
\end{figure}
\Cref{fig:riba-square-comparison-cifar10} displays the success rates of \gls*{RIBA} and Square Attack, when limiting the allowed queries to a lower number. \gls*{RIBA} can achieve a higher success rate across the board on all limits.
\subsection{ImageNet Evaluation}
\begin{figure}[tb]
    \centering
    \includegraphics[width=.47\textwidth]{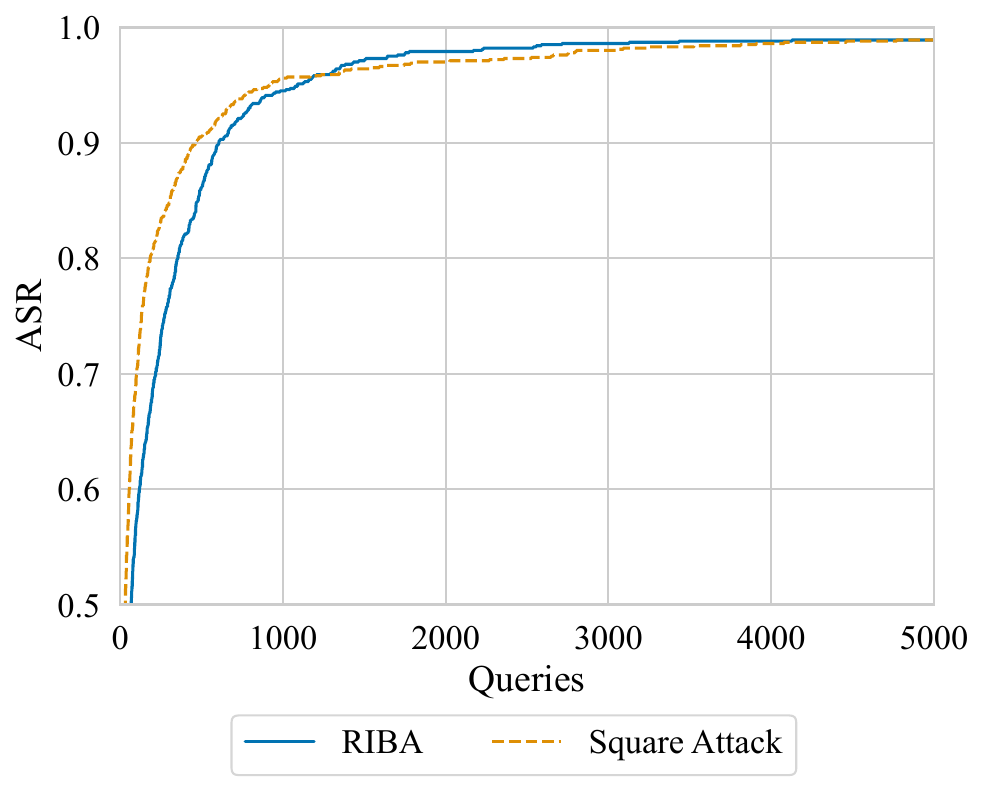}
    \caption{Comparison of the attack success rate of \gls*{RIBA} and Square Attack at different query limits when attacking an Inception-v3 model on ImageNet.}
    \label{fig:riba-square-comparison-inception}
\end{figure}

\begin{table*}[tb]
    \centering
    \caption{Experimental evaluation of attacking an Inception-v3 and a ViT-B/16 model trained on ImageNet with a perturbation limit of \(0.05\) under the \(l_\infty\) norm.}
    \label{tab:result-imagenet}
    \begin{tabular}{lrrrrrr}
    \toprule
    & \multicolumn{3}{r}{Inception-v3} & \multicolumn{3}{r}{ViT}\\
    Attack & ASR & Avg. Queries & Median Queries & ASR & Avg. Queries & Median Queries \\
    \midrule
    SimBA 
    & 84.8\%& 1567& 1021
    & 80.4\%& 1154& 690\\
    Square Attack 
    & \textbf{99.6\%} & \textbf{221} & \textbf{31} 
    & 99.9\% & \textbf{204} & 102 \\
    
    RIBA (ours) 
    & 99.2\% & 239 & 66 
    & \textbf{100\%} & 311 & \textbf{79}\\
    \bottomrule
\end{tabular}

\end{table*}
For the evaluation of our attack on ImageNet, we use an Inception-v3 model with a benign accuracy of \(77.294\%\) and a ViT-B/16 model with a benign accuracy of \(81.072\%\). We limit the perturbations to the typical budget of \(\epsilon=0.05\) and evaluate the attacks on a subset of \(1000\) randomly sampled, correctly classified images from the validation set. To reduce the computational load of generating perturbations for the higher resolution of ImageNet with our \gls*{RIBA} algorithm, we calculate the perturbations at the lower resolution of \(32\times32\) and scale them up to the image size. For the even larger images required by the Inception model of \(299\times299\), we found that it works better to first repeat the perturbation to a size of \(64\times64\) before scaling it up to the image size. This is the only hyperparameter that is tuned model, or rather input size, specific. All other hyperparameters are taken from the optimization on Cifar10. The attacks we use as a comparison are unaffected by this decision. \Cref{tab:result-imagenet} shows the attack success rate and average queries required to fool the models.
While our approach has state of the art attack success rate, it can not quite match the average queries of the highly efficient Square Attack. \Cref{fig:riba-square-comparison-inception} explores where this difference comes from in case of the Inception model. It shows the attack success rates achieved, when limiting the queries to different numbers. The Square Attack can achieve high success rates with very few queries, testimony of its highly optimized initialization. However, it falls behind for the samples harder to manipulate. For these samples, our approach is able to achieve higher success rates with fewer queries, indicating the superiority of \gls*{RIBA} for these difficult samples. This is further backed by \cref{tab:result-imagenet-hard}, which shows the average and median queries for the \(5\%\) most difficult to attack samples for Square Attack and \gls*{RIBA}. \gls*{RIBA} clearly has the edge over Square Attack, taking significantly fewer queries for difficult samples.
\begin{table}[tb]
    \centering
    \caption{Experimental evaluation of the \(5\%\) most difficult to attack samples on Inception-v3 for Square Attack and \gls*{RIBA}.}
    \label{tab:result-imagenet-hard}
    \begin{tabular}{lrr}
\toprule
Attack & Avg. Queries & Median Queries \\
\midrule
Square Attack & 2779 & 2442 \\
RIBA & 2069 & 1484 \\
\bottomrule
\end{tabular}

\end{table}
For the ViT model, our \gls*{RIBA} algorithm can significantly reduce the median queries required for a successful attack. This indicates that the initialization of Square Attack, which it heavily relies on for its query efficiency, is not efficient on transformer based models. Notably, the initialization was built specifically for convolutional networks, as vision transformers were introduced after the Square Attack was released.
\subsection{Robust Models}
To evaluate the effectiveness of \gls*{RIBA} in attacking adversarially trained models, we attack a WideResNet-82-8 trained by \citet{Bartoldson2024} on Cifar10. The model is one of the top models in the RobustBench leaderboard \citep{Croce2021}. We limit the perturbation budget to \(\epsilon=8/255\), the budget the model was trained for, and evaluate \gls*{RIBA} on \(100\) randomly sampled, correctly classified images from the Cifar10 test set. To put our results into perspective, we compare it with two white-box attacks. The simple FGSM \citep{Goodfellow2015} and the powerful PGD \citep{Madry2018a} attack.
\begin{table}[tb]
    \centering
    \caption{Experimental evaluation of attacking an adversarially trained WideResNet-82-8 on Cifar10 with a perturbation limit of \(8/255\) under the \(l_\infty\) norm.}
    \label{tab:result-robust}
    \begin{tabular}{lrrr}
    \toprule
    Attack & ASR & \makecell[r]{Avg.\\Queries} & \makecell[r]{Median\\Queries} \\
    \midrule
    FGSM & 13\% \\
    PGD & 18\% \\
    RIBA (ours) & 18\% & 1117 & 414\\
    \bottomrule
\end{tabular}

\end{table}
The PGD attack was limited to the same 10000 queries, \ie, steps, that our attack was limited to, albeit with gradient information. \Cref{tab:result-robust} shows the result of the experiment. Our \gls*{RIBA} attack was able to outperform the \acrlong{ASR} of the FGSM attack an match the PGD attack.
\subsection{Ablation Studies}
The \gls*{RIBA} algorithm is inspired by \gls*{SAC} and can be seen as a simplified, problem-specific version of it. Most of the design decisions are therefore justified by this relationship. An important design decision that was made is the reward function of the \gls*{MDP} specified in \cref{eq:reward}, especially the use of the \(\log\) function. \Cref{tab:ablation} shows the effect of this decision, as well as the effects of some hyperparameters. We generate perturbations for \(200\) randomly sampled, correctly classified images from Cifar10 targeting a ResNet-18 model and repeat the experiment three times with different seeds. One parameter at a time is altered to show its effect, while the other parameters remain at their optimized value.
\begin{table}[tb]
    \centering
    \caption{Experimental evaluation of selected design decisions and hyperparameters}
    \label{tab:ablation}
    \begin{tabular}{lrr}
\toprule
 & ASR & Avg. Queries \\
Log Reward\\
\midrule
False & 99\% & 579 \\
True & 100\% & 107 \\
\\
Starting Samples\\
\midrule
1 & 100\% & 110 \\
5 & 100\% & 107 \\
10 & 100\% & 110 \\
20 & 100\% & 115 \\
\\
Target Entropy\\
\midrule
-3072 & 100\% & 306 \\
-3 & 100\% & 107 \\
\\
Actor Update Steps\\
\midrule
20 & 100\% & 120 \\
60 & 100\% & 115 \\
120 & 100\% & 107 \\
160 & 100\% & 108 \\
\bottomrule
\end{tabular}

\end{table}
Introducing the \(\log\) function inside the reward significantly reduces the number of queries required from \(579\) to just \(107\). It amplifies small changes in the model's output and therefore helps the exploration. Increasing the number of starting samples can also help the exploration, as these randomly sampled perturbations are used in the initial training. Each sample comes at the cost of one query to the target model. Therefore, an increase in starting samples must reduce the average required queries by at least this amount. \Cref{tab:ablation} shows that for an increase from \(1\) to \(5\) starting samples, this is the case. Further increasing the number of starting samples to \(10\) does not sufficiently reduce the amount of queries during training to justify the increase. As discussed above, the target entropy is typically set to the negative number of dimensions in the action space, in our case \(-3072\). However, we found the negative number of channels in the action space, \ie, \(-3\), leads to better results. This is backed by \cref{tab:ablation}, showing that it requires less than half of the average queries. Finally, we look into the effect of increasing the actor update steps. This hyperparameter specifies how many times the policy is updated per query to the target model. It is clearly visible that, up to a certain point, an increase in update steps can significantly reduce the average number of queries, albeit with diminishing returns. However, too large values can lead to an increase in queries, as the actor is trained on increasingly outdated information.

\subsection{Batched Inputs}
When attacking a model in a real world use case, an attacker might not be interested in generating perturbations against thousands, but rather against very few select images. In this section, we explore a unique possibility of our approach to further reduce the amount of times the target model is queried, when generating a perturbation for a single image. One of the major features of off-policy \gls*{RL} is the separation of the collection of experience from the training. Our \gls*{RIBA} algorithm follows the same paradigm. In \cref{alg:riba}, perturbations are sampled and evaluated and then added to the replay buffer starting at Line \ref{alg-line:sample-perturbation-start}. This requires querying the target model. Afterwards, the actor and critic are trained using past experiences from the replay buffer starting at Line \ref{alg-line:training-start}.
In \gls*{SAC}, it is possible to execute multiple steps in the environment and add each transition to the replay buffer, before training again \citep{Haarnoja2018}. The same is possible in \gls*{RIBA}. However, in our case, the next perturbation is not dependent on the previous one, like it would be for a typical \gls*{MDP}. Therefore, we can generate multiple perturbations at once and query the target model in one batch.
To investigate the effect of trying multiple perturbations at once, we utilize the same setup as for the ablation studies, \ie, we generate perturbations for \(200\) randomly sampled, correctly classified images from Cifar10, targeting a ResNet-18 model, and repeating the experiment with three different seeds.
\begin{figure}
    \centering
    \includegraphics[width=.47\textwidth]{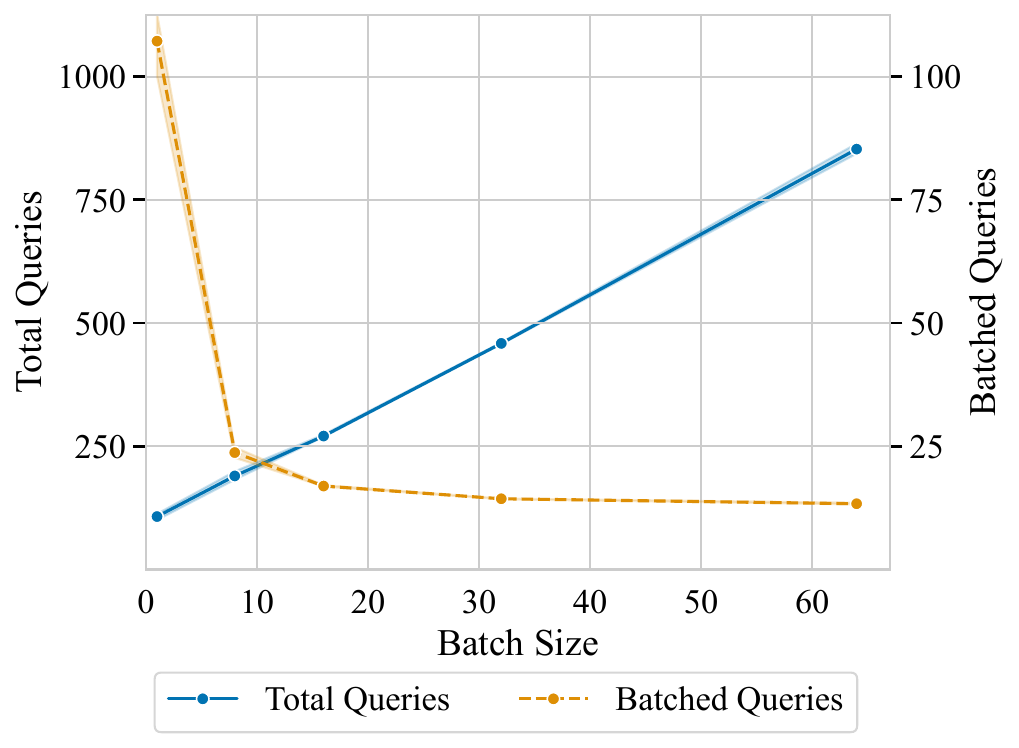}
    \caption{Visualization of how batched queries to the target model influence the amount of queries needed in total, as well as the amount of batches.}
    \label{fig:env-batch-size}
\end{figure}
\Cref{fig:env-batch-size} shows the effect of increasing the batch size used to query the target model from \(1\) up to \(64\). The total amount of queries required increases linearly with an increased batch size. However, the amount of batches, \ie, the number of times, the target model is actually called, can be significantly reduced. Increasing the batch size from \(1\) to \(8\) can reduce the number of batched queries from \(107\) to \(24\). Further increasing the batch size reduces the number of batched queries even further. However, we observe diminishing returns with larger batch sizes, albeit with only \(13\) batched queries required at a batch size of \(64\). This shows a significant advantage of \gls*{RIBA} in a realistic use case over other attacks that do not support batched queries when attacking a single image.

\section{Discussion}
\label{sec:discussion}
We successfully demonstrated the capabilities of our novel approach in the previous section. Specifically, we showed that our approach can attack different models on different data sets, while only requiring a small number of queries to the target model.

We were able to show that our approach performs similarly to the long standing state of the art Square Attack. \gls*{RIBA} clearly outperforms Square Attack on the Resnet18 model trained on Cifar10. For the more difficult case of models trained on ImageNet, the results are not as clear. Due to its initialization, which is highly optimized for fooling convolutional networks, Square Attack can generate perturbations with fewer queries than \gls*{RIBA} on most samples against the Inception-v3 model. However, \gls*{RIBA} performs significantly better on the more difficult samples. On the transformer based ViT-B/16 model, \gls*{RIBA} takes significantly fewer median queries than Square Attack, indicating that the initialization of Square Attack does not work well on transformer based architectures. Both attacks have their strength in different areas. Notably, they follow entirely different paradigms, an \gls*{RL} inspired optimization and random search. It is of utmost importance that different attack strategies are known to the research community, in order to properly defend models against all types of attacks.

We designed the \gls*{RIBA} algorithm for the \(l_\infty\) threat model. However, an adaptation to the \(l_2\) threat model should be possible. The \(l_\infty\)-norm bound is enforced statically due to the way perturbations are applied to the images in \cref{eq:apply-perturbation}. This is not possible for the \(l_2\)-norm bound. However, the \(l_2\)-norm of the perturbation could be added as a penalty to the reward function, or directly added to the training objective of the policy in \cref{eq:policy-objective} as a penalty. We leave it for future work to explore these possibilities.

\section{Conclusion}
\label{sec:conclusion}
In this paper, we introduced a novel approach to optimize adversarial perturbations under the realistic back-box threat model. First, we formulated the problem of optimizing adversarial perturbations as a \acrlong{MDP}, which would allow the use of any \acrlong{RL} algorithm to optimize perturbations. 
Then, we developed the novel \acrfull{RIBA} algorithm, tailored to specifically solve this MDP. \gls*{RIBA} is based  on the concepts of the \acrfull{SAC} algorithm.
We showed the success of our attack by comparing it to the long standing state of the art Square Attack. \gls*{RIBA} requires significantly fewer queries to generate perturbations on the Cifar10 data set to fool a ResNet-18 model than the Square Attack. Additionally, it requires fewer median queries to fool the transformer based ViT-B/16 on ImageNet. For the convolutional Inception-v3 model, we were able to show that \gls*{RIBA} requires fewer queries to generate perturbations for particularly difficult to manipulate images. On an adversarially trained model, we were able to match the attack success rate of the white-box PGD attack. Finally, we showed that with \gls*{RIBA}, queries can be grouped together into batches, even when generating a perturbation for a single image, further reducing the amount of times the target model is called. This is a unique property of \gls*{RIBA} that stems from the inspiration from off-policy \gls*{RL}.

\section{Ethics statement}
\label{sec:ethics}
We are aware of the significant risk that adversarial attacks pose to the safety of computer vision systems applied in the real world. Black-box attacks, such as the one presented in this paper, are especially dangerous, as they are able to attack systems which can only be accessed through an API.
We release this novel approach to outline the possibility of the attack and spark the creation of tailored defense mechanisms. It is important that novel attacks are discovered through red teaming, giving machine learning practitioners the chance to mitigate them, before the attacks are employed by malicious actors.

{
    \small
    \bibliographystyle{ieeenat_fullname}
    \bibliography{references}

\begin{thebibliography}{30}
\providecommand{\natexlab}[1]{#1}
\providecommand{\url}[1]{\texttt{#1}}
\expandafter\ifx\csname urlstyle\endcsname\relax
  \providecommand{\doi}[1]{doi: #1}\else
  \providecommand{\doi}{doi: \begingroup \urlstyle{rm}\Url}\fi

\bibitem[Andriushchenko et~al.(2020)Andriushchenko, Croce, Flammarion, and
  Hein]{Andriushchenko2020}
Maksym Andriushchenko, Francesco Croce, Nicolas Flammarion, and Matthias Hein.
\newblock Square {Attack}: {A} {Query}-{Efficient} {Black}-{Box} {Adversarial}
  {Attack} via {Random} {Search}.
\newblock In \emph{Computer {Vision} – {ECCV} 2020}, pages 484--501, Cham,
  2020. Springer International Publishing.

\bibitem[Bai et~al.(2023)Bai, Wang, Zeng, Jiang, and Xia]{Bai2023}
Yang Bai, Yisen Wang, Yuyuan Zeng, Yong Jiang, and Shu-Tao Xia.
\newblock Query efficient black-box adversarial attack on deep neural networks.
\newblock \emph{Pattern Recognition}, 133:\penalty0 109037, 2023.

\bibitem[Bartoldson et~al.(2024)Bartoldson, Diffenderfer, Parasyris, and
  Kailkhura]{Bartoldson2024}
Brian~R. Bartoldson, James Diffenderfer, Konstantinos Parasyris, and Bhavya
  Kailkhura.
\newblock Adversarial {Robustness} {Limits} via {Scaling}-{Law} and
  {Human}-{Alignment} {Studies}.
\newblock In \emph{Proc. {Mach}. {Learn}. {Res}.}, pages 3046--3072. ML
  Research Press, 2024.

\bibitem[Carlini and Wagner(2017)]{Carlini2017}
Nicholas Carlini and David Wagner.
\newblock Towards {Evaluating} the {Robustness} of {Neural} {Networks}.
\newblock In \emph{2017 IEEE Symposium on Security and Privacy (SP)}, pages
  39--57. IEEE, 2017.
\newblock ISSN: 2375-1207.

\bibitem[Chen et~al.(2017)Chen, Zhang, Sharma, Yi, and Hsieh]{Chen2017}
Pin-Yu Chen, Huan Zhang, Yash Sharma, Jinfeng Yi, and Cho-Jui Hsieh.
\newblock {ZOO}: {Zeroth} {Order} {Optimization} {Based} {Black}-box {Attacks}
  to {Deep} {Neural} {Networks} without {Training} {Substitute} {Models}.
\newblock In \emph{Proceedings of the 10th {ACM} {Workshop} on {Artificial}
  {Intelligence} and {Security}}, pages 15--26, New York, NY, USA, 2017.
  Association for Computing Machinery.

\bibitem[Chen et~al.(2019)Chen, Liu, Xu, Li, Lin, Hong, and Cox]{Chen2019}
Xiangyi Chen, Sijia Liu, Kaidi Xu, Xingguo Li, Xue Lin, Mingyi Hong, and David
  Cox.
\newblock {ZO}-{AdaMM}: {Zeroth}-{Order} {Adaptive} {Momentum} {Method} for
  {Black}-{Box} {Optimization}.
\newblock In \emph{Advances in {Neural} {Information} {Processing} {Systems}}.
  Curran Associates, Inc., 2019.

\bibitem[Croce et~al.(2021)Croce, Andriushchenko, Sehwag, Debenedetti,
  Flammarion, Chiang, Mittal, and Hein]{Croce2021}
Francesco Croce, Maksym Andriushchenko, Vikash Sehwag, Edoardo Debenedetti,
  Nicolas Flammarion, Mung Chiang, Prateek Mittal, and Matthias Hein.
\newblock {RobustBench}: a standardized adversarial robustness benchmark.
\newblock In \emph{Adv. neural inf. proces. syst.} Neural information
  processing systems foundation, 2021.

\bibitem[Dosovitskiy et~al.(2021)Dosovitskiy, Beyer, Kolesnikov, Weissenborn,
  Zhai, Unterthiner, Dehghani, Minderer, Heigold, Gelly, Uszkoreit, and
  Houlsby]{Dosovitskiy2021}
Alexey Dosovitskiy, Lucas Beyer, Alexander Kolesnikov, Dirk Weissenborn,
  Xiaohua Zhai, Thomas Unterthiner, Mostafa Dehghani, Matthias Minderer, Georg
  Heigold, Sylvain Gelly, Jakob Uszkoreit, and Neil Houlsby.
\newblock An image is worth 16x16 words: Transformers for image recognition at
  scale.
\newblock In \emph{{ICLR} - {Int}. {Conf}. {Learn}. {Represent}.} International
  Conference on Learning Representations, ICLR, 2021.

\bibitem[Goodfellow et~al.(2015)Goodfellow, Shlens, and
  Szegedy]{Goodfellow2015}
Ian~J. Goodfellow, Jonathon Shlens, and Christian Szegedy.
\newblock Explaining and {Harnessing} {Adversarial} {Examples}, 2015.
\newblock arXiv:1412.6572 [stat].

\bibitem[Guo et~al.(2019)Guo, Gardner, You, Wilson, and Weinberger]{Guo2019}
Chuan Guo, Jacob Gardner, Yurong You, Andrew~Gordon Wilson, and Kilian
  Weinberger.
\newblock Simple {Black}-box {Adversarial} {Attacks}.
\newblock In \emph{Proceedings of the 36th International Conference on Machine
  Learning}, pages 2484--2493. PMLR, 2019.

\bibitem[Haarnoja et~al.(2018)Haarnoja, Zhou, Abbeel, and Levine]{Haarnoja2018}
Tuomas Haarnoja, Aurick Zhou, Pieter Abbeel, and Sergey Levine.
\newblock Soft {Actor}-{Critic}: {Off}-{Policy} {Maximum} {Entropy} {Deep}
  {Reinforcement} {Learning} with a {Stochastic} {Actor}.
\newblock In \emph{Proceedings of the 35th International Conference on Machine
  Learning}, pages 1861--1870. PMLR, 2018.

\bibitem[Haarnoja et~al.(2019)Haarnoja, Zhou, Hartikainen, Tucker, Ha, Tan,
  Kumar, Zhu, Gupta, Abbeel, and Levine]{Haarnoja2019}
Tuomas Haarnoja, Aurick Zhou, Kristian Hartikainen, George Tucker, Sehoon Ha,
  Jie Tan, Vikash Kumar, Henry Zhu, Abhishek Gupta, Pieter Abbeel, and Sergey
  Levine.
\newblock Soft {Actor}-{Critic} {Algorithms} and {Applications}, 2019.
\newblock arXiv:1812.05905 [cs.LG].

\bibitem[He et~al.(2016)He, Zhang, Ren, and Sun]{He2016}
Kaiming He, Xiangyu Zhang, Shaoqing Ren, and Jian Sun.
\newblock Deep residual learning for image recognition.
\newblock In \emph{Proc {IEEE} {Comput} {Soc} {Conf} {Comput} {Vision}
  {Pattern} {Recognit}}, pages 770--778. IEEE Computer Society, 2016.

\bibitem[Huang et~al.(2023)Huang, Zhou, Hefenbrock, Riedel, Fang, and
  Beigl]{Huang2023}
Yiran Huang, Yexu Zhou, Michael Hefenbrock, Till Riedel, Likun Fang, and
  Michael Beigl.
\newblock Universal {Distributional} {Decision}-{Based} {Black}-{Box}
  {Adversarial} {Attack} with {Reinforcement} {Learning}.
\newblock In \emph{Lect. {Notes} {Comput}. {Sci}.}, pages 206--215. Springer
  Science and Business Media Deutschland GmbH, 2023.

\bibitem[Ilyas et~al.(2018)Ilyas, Engstrom, Athalye, and Lin]{Ilyas2018}
Andrew Ilyas, Logan Engstrom, Anish Athalye, and Jessy Lin.
\newblock Black-box {Adversarial} {Attacks} with {Limited} {Queries} and
  {Information}.
\newblock In \emph{Proceedings of the 35th International Conference on Machine
  Learning}, pages 2137--2146. PMLR, 2018.

\bibitem[Ilyas et~al.(2019)Ilyas, Engstrom, and Madry]{Ilyas2019}
Andrew Ilyas, Logan Engstrom, and Aleksander Madry.
\newblock Prior convictions: Black-box adversarial attacks with bandits and
  priors.
\newblock In \emph{International Conference on Learning Representations}, 2019.

\bibitem[Kang et~al.(2023)Kang, Song, Guo, Qin, Du, and Guizani]{Kang2023}
Xu Kang, Bin Song, Jie Guo, Hao Qin, Xiaojiang Du, and Mohsen Guizani.
\newblock Black-box attacks on image classification model with advantage
  actor-critic algorithm in latent space.
\newblock \emph{Information Sciences}, 624:\penalty0 624--638, 2023.

\bibitem[Krizhevsky et~al.(2009)]{Krizhevsky2009}
Alex Krizhevsky et~al.
\newblock Learning multiple layers of features from tiny images.
\newblock 2009.

\bibitem[Loshchilov and Hutter(2019)]{Loshchilov2019}
Ilya Loshchilov and Frank Hutter.
\newblock Decoupled weight decay regularization.
\newblock In \emph{Int. {Conf}. {Learn}. {Represent}., {ICLR}}. International
  Conference on Learning Representations, ICLR, 2019.

\bibitem[Ma et~al.(2021)Ma, Chen, and Yong]{Ma2021}
Chen Ma, Li Chen, and Jun-Hai Yong.
\newblock Simulating unknown target models for query-efficient black-box
  attacks.
\newblock In \emph{Proceedings of the IEEE/CVF Conference on Computer Vision
  and Pattern Recognition (CVPR)}, pages 11835--11844, 2021.

\bibitem[Ma and Feng(2025)]{Ma2025}
Zerou Ma and Tao Feng.
\newblock Query-{Efficient} {Two}-{Phase} {Reinforcement} {Learning}
  {Framework} for {Black}-{Box} {Adversarial} {Attacks}.
\newblock \emph{Symmetry}, 17\penalty0 (7):\penalty0 1093, 2025.

\bibitem[Madry et~al.(2018)Madry, Makelov, Schmidt, Tsipras, and
  Vladu]{Madry2018a}
Aleksander Madry, Aleksandar Makelov, Ludwig Schmidt, Dimitris Tsipras, and
  Adrian Vladu.
\newblock Towards deep learning models resistant to adversarial attacks.
\newblock In \emph{Int. {Conf}. {Learn}. {Represent}., {ICLR} - {Conf}. {Track}
  {Proc}.} International Conference on Learning Representations, ICLR, 2018.

\bibitem[Russakovsky et~al.(2015)Russakovsky, Deng, Su, Krause, Satheesh, Ma,
  Huang, Karpathy, Khosla, Bernstein, Berg, and Fei-Fei]{Russakovsky2015}
Olga Russakovsky, Jia Deng, Hao Su, Jonathan Krause, Sanjeev Satheesh, Sean Ma,
  Zhiheng Huang, Andrej Karpathy, Aditya Khosla, Michael Bernstein,
  Alexander~C. Berg, and Li Fei-Fei.
\newblock {ImageNet} {Large} {Scale} {Visual} {Recognition} {Challenge}.
\newblock \emph{International Journal of Computer Vision}, 115\penalty0
  (3):\penalty0 211--252, 2015.

\bibitem[Sarkar et~al.(2023)Sarkar, Babu, Mousavi, Ghorbanpour, Gundecha,
  Guillen, Luna, and Naug]{Sarkar2023a}
Soumyendu Sarkar, Ashwin~Ramesh Babu, Sajad Mousavi, Sahand Ghorbanpour, Vineet
  Gundecha, Antonio Guillen, Ricardo Luna, and Avisek Naug.
\newblock Robustness with query-efficient adversarial attack using
  reinforcement learning.
\newblock In \emph{Proceedings of the IEEE/CVF Conference on Computer Vision
  and Pattern Recognition (CVPR) Workshops}, pages 2330--2337, 2023.

\bibitem[Szegedy et~al.(2014)Szegedy, Zaremba, Sutskever, Bruna, Erhan,
  Goodfellow, and Fergus]{Szegedy2014}
Christian Szegedy, Wojciech Zaremba, Ilya Sutskever, Joan Bruna, Dumitru Erhan,
  Ian Goodfellow, and Rob Fergus.
\newblock Intriguing properties of neural networks, 2014.
\newblock arXiv:1312.6199 [cs].

\bibitem[Szegedy et~al.(2016)Szegedy, Vanhoucke, Ioffe, Shlens, and
  Wojna]{Szegedy2016}
Christian Szegedy, Vincent Vanhoucke, Sergey Ioffe, Jon Shlens, and Zbigniew
  Wojna.
\newblock Rethinking the {Inception} {Architecture} for {Computer} {Vision}.
\newblock In \emph{Proc {IEEE} {Comput} {Soc} {Conf} {Comput} {Vision}
  {Pattern} {Recognit}}, pages 2818--2826. IEEE Computer Society, 2016.

\bibitem[Xiong et~al.(2022)Xiong, Lin, Zhang, Hopcroft, and He]{Xiong2022}
Yifeng Xiong, Jiadong Lin, Min Zhang, John~E. Hopcroft, and Kun He.
\newblock Stochastic {Variance} {Reduced} {Ensemble} {Adversarial} {Attack} for
  {Boosting} the {Adversarial} {Transferability}.
\newblock In \emph{Proc {IEEE} {Comput} {Soc} {Conf} {Comput} {Vision}
  {Pattern} {Recognit}}, pages 14963--14972. IEEE Computer Society, 2022.

\bibitem[Yang et~al.(2020)Yang, Jiang, Huang, Ni, and Zhao]{YANG2020}
Jiancheng Yang, Yangzhou Jiang, Xiaoyang Huang, Bingbing Ni, and Chenglong
  Zhao.
\newblock Learning {Black}-{Box} {Attackers} with {Transferable} {Priors} and
  {Query} {Feedback}.
\newblock In \emph{Advances in {Neural} {Information} {Processing} {Systems}},
  pages 12288--12299. Curran Associates, Inc., 2020.

\bibitem[Zhao et~al.(2020)Zhao, Chen, Wang, and Lin]{Zhao2020}
Pu Zhao, Pin-yu Chen, Siyue Wang, and Xue Lin.
\newblock Towards {Query}-{Efficient} {Black}-{Box} {Adversary} with
  {Zeroth}-{Order} {Natural} {Gradient} {Descent}.
\newblock \emph{Proceedings of the AAAI Conference on Artificial Intelligence},
  34\penalty0 (04):\penalty0 6909--6916, 2020.

\bibitem[Zhou et~al.(2022)Zhou, Cui, Zhang, Jiang, and Yang]{Zhou2022}
Linjun Zhou, Peng Cui, Xingxuan Zhang, Yinan Jiang, and Shiqiang Yang.
\newblock Adversarial eigen attack on black-box models.
\newblock In \emph{Proceedings of the IEEE/CVF Conference on Computer Vision
  and Pattern Recognition (CVPR)}, pages 15254--15262, 2022.

\end{thebibliography}
}

\end{document}